%% file: main.tex
\documentclass[runningheads]{llncs}

\usepackage[final]{eccv}
\usepackage{eccvabbrv}

\usepackage{graphicx}
\usepackage{booktabs}
\usepackage{url}
\usepackage{graphicx}
\usepackage{float}
\usepackage{multirow}
\usepackage{booktabs}
\usepackage{algorithm}
\usepackage{multicol,multirow}
\usepackage{algpseudocode}
\usepackage{adjustbox}
\usepackage{wrapfig}
\usepackage[accsupp]{axessibility}  % Improves PDF readability for those with disabilities.

\def\bx{{\mathbf x}}

\usepackage[pagebackref,breaklinks,colorlinks,citecolor=eccvblue]{hyperref}
\usepackage{orcidlink}

\begin{document}

\title{Streaming Video Diffusion: Online Video Editing with Diffusion Models} 

\titlerunning{SVDiff}

\author{Feng Chen$^{1 \dagger}$ \and
Zhen Yang$^{2 \dagger}$ \and
Bohan Zhuang$^{3 \ddagger}$
\and
Qi Wu$^{1 \ddagger}$
}

\authorrunning{Feng Chen et al.}

\institute{Australian Institute for Machine Learning, The University of Adelaide \and
Zhejiang University
\\
\and
Monash University\\
}

\maketitle

\input{sec/0_abstract}    

\newcommand\blfootnote[1]{% 
\begingroup 
\renewcommand\thefootnote{}\footnote{#1}% 
\addtocounter{footnote}{-1}% 
\endgroup 
}
{
	\blfootnote{
	 %{{$^\scriptscriptstyle \spadesuit$}} 
  $^\dagger$ Co-first authors.\\
  ~$^\ddagger$ Corresponding authors: \href{mailto:bohan.zhuang@gmail.com}{\color{black}{bohan.zhuang@gmail.com}}, \href{mailto:qi.wu01@adelaide.edu.au}{\color{black}{qi.wu01@adelaide.edu.au}}.
	}
}
\input{sec/1_intro}
\input{sec/2_related}

\input{sec/3_method}

\input{sec/4_result}

\input{sec/5_conclusion}

%\clearpage  % TODO REVIEW/FINAL: This \clearpage needs to be removed from both review and camera-ready versions.

\bibliographystyle{splncs04}
\bibliography{main}
\end{document}

%% file: sec/0_abstract.tex
\begin{abstract}
We present a novel task called online video editing, which is designed to edit \textbf{streaming} frames while maintaining temporal consistency. Unlike existing offline video editing assuming all frames are pre-established and accessible, online video editing is tailored to real-life applications such as live streaming and online chat, requiring (1) fast continual step inference, (2) long-term temporal modeling, and (3) zero-shot video editing capability. To solve these issues, we propose Streaming Video Diffusion (SVDiff), which incorporates the compact spatial-aware temporal recurrence into off-the-shelf Stable Diffusion and is trained with the segment-level scheme on large-scale long videos. This simple yet effective setup allows us to obtain a single model that is capable of executing a broad range of videos and editing each streaming frame with temporal coherence. Our experiments indicate that our model can edit long, high-quality videos with remarkable results, achieving a real-time inference speed of 15.2 FPS at a resolution of $512 \times 512$.
Our code will be available at \url{https://github.com/Chenfeng1271/SVDiff}.

  \keywords{Video editing \and Streaming processing \and Diffusion}

\end{abstract}

%% file: sec/1_intro.tex
\section{Introduction}
\label{sec:intro}
\begin{figure}[tbp]
\centering
\includegraphics[width=0.85\linewidth]{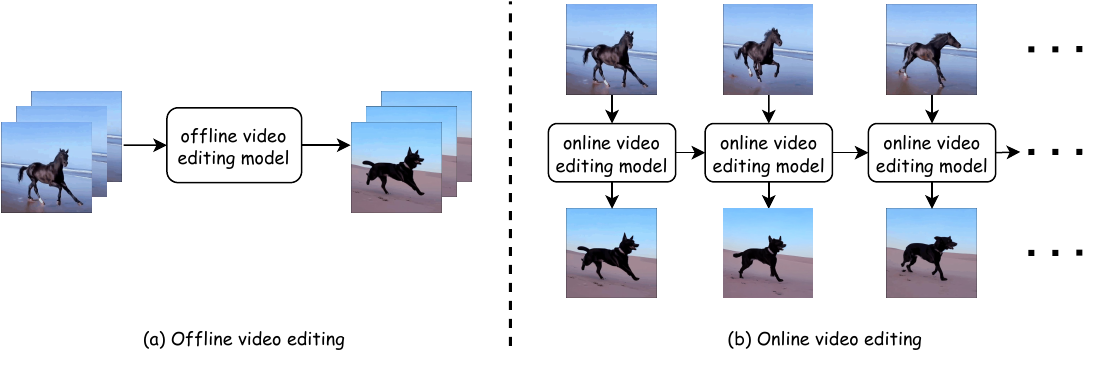}\caption{Comparison between offline and online video editing. Offline video editing processes the whole video simultaneously and regards all frames as known. Online video editing operates each streaming frame with the temporal information from previous frames in a causal way.}
\label{figure task}
\end{figure}

Video editing~\cite{wu2022tuneavideo, he2022lvdm,zhang2023magicbrush} plays a ubiquitous role in creating fascinating visual effects for films, short videos, \etc. 
Recent advancements~\cite{ceylan2023pix2video,Dreamix} have predominantly concentrated on offline video editing (as shown in \cref{figure task} (a)), wherein the entire video is edited simultaneously, assuming that all frames are pre-established and accessible.
However, as shown in \cref{figure task} (b), editing streaming frames of video for immediate response to visual data, which we call \textit{online video editing}, is still underexplored. 
It is important for many real-life usage scenarios such as live streaming and online chat. 
%\textcolor{blue}{Unlike online video recognition~~\cite{zhao2023streaming} that typically identifies and categorizes task-related content within the stream, online video editing necessitates the temporal alignment of motion trajectory across successive frames.}
As a result, there is an increasing demand for easy-to-use and performant online video editing tools.

Recently, thanks to the introduction of powerful text-conditioned diffusion models~\cite{ramesh2022dalle2, imagen-video, rombach2022high} trained on large-scale datasets, video editing algorithms have achieved unprecedented processes in offline video editing by extending Text-to-Image (T2I) to Text-to-Video (T2V) diffusion. 
Typically, sparse causal attention ~\cite{ceylan2023pix2video,wang2018vid2vid,qi2023fatezero} and temporal module ~\cite{wu2022tuneavideo, ma2023follow,chen2023videocrafter1} are added to model temporal dynamics, but they are insufficient for online video editing due to short-term temporal modeling and accessing future frames, respectively~\cite{genlvideo,harvey2022flexible}. 
%\bohan{add more references}
Therefore, it is still challenging to extend this success to online video editing. 
We summarize these challenges as three-fold.
1): The multi-step denoising of diffusion significantly increases the computational redundancy of cached memory and recurrent calculation, which is difficult for fast continual inference.
2): Online video streams usually have an extended video sequence, which requires long-term temporal modeling. However, training a model on long videos is non-trivial.
3): For online video editing to be both practical and effective, each model must possess zero-shot video editing capabilities, allowing the editing of any video in response to any edit prompt.

To address these issues, one straightforward approach is to adapt existing zero-shot offline methods to the causal online setting. 
%, but it is still problematic.
%\bohan{put reference here}
%\bohan{why suddenly mention text-to-video?? are you doing video generation tasks? also, can you follow the "categories names" of the video editing literature?}
These zero-shot offline methods can be classified into two types: tuning-free based and pretrained-based.
Tuning-free based methods~\cite{wu2022nuwa, ceylan2023pix2video} apply additional controls (such as replacing spatial attention with sparse causal attention) to maintain frame-to-frame consistency. 
These adjustments primarily preserve temporal coherence over short spans~\cite{zhao2023streaming}, which makes them suitable for editing brief videos, but less effective for longer sequences.
%\bohan{why within a few frames?}
To adapt these tuning-free methods for online video editing, it is essential to use all previous frames for cross-frame interaction. 
However, the amount of data involved is substantial, especially in videos of high resolution and long duration, leading to a drastic increase in memory consumption~\cite{harvey2022flexible}.
%\bohan{wrong. This is very efficient}
%\bohan{??}
%\bohan{first setting. then highlight adapting challenges..refer to Figure 1 (a), (b), (c)}
The other type of offline method called pretrained-based methods~\cite{ma2023follow, singer2023makeavideo}, turns to training the video diffusion model on a large-scale text-video dataset to model temporal dynamics. 
Typically, temporal attention ~\cite{chen2023videocrafter1} is added to the denoising model, fostering inter-frame interaction. 
The parallel calculation of the attention module regards all frames as known, which is contradictory with online video editing because a future frame is not available at the current time. 
A simple way to adapt these pretrained-based methods to online video editing is to add causal attention masks to their temporal attention. 
However, this approach involves recalculating previous frames for every new frame in the stream, a process that is ill-suited for rapid inference. Recently, LLaVA~\cite{llava} introduced a method that caches previous key and value states to bypass repetitive computations. The drawback of this strategy is the increased memory requirement for storage, particularly noticeable during multi-step denoising.

In this paper, we propose an online video diffusion model with recursive spatial-aware temporal memory, named streaming video diffusion (SVDiff), to balance the trade-off between
computational cost and long-range temporal modeling. 
%\bohan{as shown in Figure 1 (d).}
Specifically, as shown in \cref{fig:overview}, we first initialize a learnable spatial-aware temporal memory embedding and recursively process it with streaming frames. 
It essentially serves as a dynamic temporal cache and is continuously updated by memory attention to encode both the individual content of each frame's spatial layout and the inter-frame motion trajectory within the video stream.
Therefore, it allows for trivial computational cost %as $\mathcal{O}(H^2W^2)$ 
and long-range temporal modeling for online processing, resulting from compact memory and recurrent operation. 
After that, we adopt the segment-level scheme~\cite{genlvideo} that deconstructs a long video into a series of short video clips for efficient long video training. 
Apart from updating and processing memory within each segmented clip, we also propagate the temporal memory between consecutive clips, transferring the temporal history to the following video frames. 
Unlike previous methods that usually edit 16-frame videos, this setup allows us to train a single model that is capable of executing videos with 150 frames and editing each streaming frame with temporal consistency.

To sum up, we make three main contributions: 1) We propose online video editing, a novel task for immediate editing response of streaming video. 2) We propose SVDiff, an online video diffusion model with recursive spatial-aware temporal memory. 3) Our method efficiently generates high-quality, long videos, ensuring both global and local coherence, while maintaining a real-time inference speed of 15.2 FPS with a resolution of $512 \times 512$.

%% file: sec/2_related.tex
\section{Related Work}

%\bohan{don't simply summarize what xxx does...you need to analyze what are their drawbacks...Contrary to xxx, our method xxxx}
%\bohan{Are you writing pretrained video generation?}
\noindent\textbf{Pretrained-based video editing.} Despite considerable progress in zero-shot text-guided image editing ~\cite{hertz2022prompt}, editing arbitrary videos according to text remains a difficult task due to the lack of large-scale high-quality text-video datasets and the complexity of modeling temporal consistency. 
To solve this issue, Dreamix~\cite{Dreamix} and LAMP~\cite{wu2023lamp} propose to train a video diffusion model over T2I diffusion on a small dataset for video editing. 
In addition, FollowYourPose~\cite{ma2023follow} learns a separate pose branch to maintain the structure of the video. Recently, Videocrafter1 ~\cite{chen2023videocrafter1} introduced a high-resolution video diffusion pre-trained on a large-scale text-video dataset, which can preserve content, style and motion consistency. These methods usually employ temporal modules, such as attention~\cite{chen2023videocrafter1}, LoRA~\cite{ma2023follow}, and convolution~\cite{singer2023makeavideo}, to capture temporal changes, but these modules treat all frames as known, which is inconsistent with online video editing. 
%\bohan{what's v-prediction parameterization?? you need to make it readable for general audience...}

\noindent\textbf{Tuning-free video editing.} Another way for zero-shot video editing is to modify T2I diffusion in a tuning-free design.
For example, to achieve temporal consistency, Fate-Zero~\cite{qi2023fatezero} and Video-P2P~\cite{liu2023video} replace spatial attention in U-Net with sparse causal attention and apply attention control proposed in prompt2prompt~\cite{hertz2022prompt}. 
Pix2Video~\cite{ceylan2023pix2video} adds additional regularization to penalize dramatic frame changes. 
Text2Video-Zero~\cite{khachatryan2023text2video} first proposes to edit the video frames with only the pre-trained T2I diffusion model and then modify the latent feature with motion dynamic through sparse causal attention and object mask. However, because only a few frames used in cross-frame attention, these methods still struggle to maintain long-range temporal consistency.

\noindent\textbf{Online video models.} Online video processing refers to the analysis and manipulation of video content as it is being streamed or captured, enabling immediate interpretation and response to visual data. Existing methods mainly focus on online video recognition with recursive operation~\cite{yang2022recurring}, temporal shift~\cite{liu2022ts2}, sliding window~\cite{ai2022class}, and augmented memory \cite{zhao2023streaming}. For example, Yang et al. ~\cite{yang2022recurring} use a recurrent attention gate to aggregate the information between the current frame and previous frames. \cite{zhao2023streaming} caches the key-value of previous frames, acting as a temporal reference in cross-frame attention. ~\cite{liu2022ts2} selects the informative tokens from each frame and then temporally shifts them across the adjacent frames. However, in online video editing with diffusion models, multi-step diffusion denoising poses significant challenges in modeling long-term motion trajectories and achieving fast inference. In this paper, we propose a novel method using compact temporal recurrence to solve this issue. %which is demonstrated to outperform baseline models adapting previous methods to online video editing.
%\bohan{???}

%% file: sec/3_method.tex
\section{Preliminary}

Stable Diffusion~\cite{rombach2022high, podell2023sdxl} is a latent diffusion model operating within the latent space of an autoencoder. 
We denote the autoencoder as $\mathcal{D}(\mathcal{E}(\cdot))$, where $\mathcal{E}$ and $\mathcal{D}$ correspond to the encoder and decoder, respectively.
Taking an input image $I$ and its corresponding latent feature $\bx_0$ obtained through the encoder $\bx_0 = \mathcal{E}(I)$, the core of the diffusion process is to iteratively introduce noise to this latent representation. This is achieved through the following equation:
%For an image $I$ with its latent feature $\bx_0=\mathcal{E}(I)$, the diffusion forward process iteratively introduces noise to the latent representation:
\begin{equation} 
q(\bx_t\mid\bx_{t-1})=\mathcal{N}(\bx_t;\sqrt{\alpha_t}\bx_{t-1}, (1-\alpha_t)\mathbf{I}), 
\end{equation} 
where $t=1,...,T$ is the time step, $q(\bx_t\mid\bx_{t-1})$ is the conditional density of $\bx_t$ given $\bx_{t-1}$, and $\alpha_t$ is a hyperparameter to scale noise. Alternatively, at any given time step, $\bx_t$ can be sampled from $\bx_0$ using the following equation:
\begin{equation}
\label{eq:forward_sample} 
q(\bx_t\mid\bx_{0})=\mathcal{N}(\bx_t;\sqrt{\bar{\alpha}_t}\bx_0, (1-\bar{\alpha}_t)\mathbf{I}), 
\end{equation} 
where $\bar{\alpha}_t=\prod_{i=1}^t\alpha_i$. In the diffusion backward process, a U-Net denoted as $\epsilon_\theta$ is trained to predict the noise in the latent representation, aiming to iteratively recover $\bx_0$ from $\bx_T$. 
As the number of diffusion steps, denoted as $T$, increases, $\bx_0$ becomes progressively noisier due to the noise introduced in the forward process.
This noise accumulation causes $\bx_T$ to approximate a standard Gaussian distribution.
Consequently, $\epsilon_\theta$ is designed to learn how to deduce a valid $\bx_0$ from these Gaussian noises.
% After training, we can sample $\bx_{t-1}$ from $\bx_{t}$ using DDIM sampler:
% \begin{equation}\label{eq:ddim}
%   \bx_{t-1}=\sqrt{\alpha_{t-1}}\underbrace{\hat{\bx}_{t\rightarrow0}}_{\text{predicted}~x_0}
%   +\underbrace{\sqrt{1-\alpha_{t-1}}\epsilon_\theta(\bx_t, t, c_p)}_{\text{direction pointing to}~\bx_{t-1}},
% \end{equation}
Given $c_p$ is the text prompt, the predicted $\bx_0$, denoted as $\hat{\bx}_{t\rightarrow0}$ at time step $t$, can be estimated using the following equation:
\begin{equation}
  \hat{\bx}_{t\rightarrow0}=(\bx_t-\sqrt{1-\alpha_{t}}\epsilon_\theta(\bx_t, t, c_p))/\sqrt{\alpha_{t}},
\end{equation}
where $\epsilon_\theta(\bx_t, t, c_p)$ is the predicted noise of $\bx_t$ guided by the text prompt $c_p$ and the time step $t$. 
Meanwhile, the reconstruction loss between the real noise $\epsilon_t$ and $\epsilon_{\theta}(\bx_t,t,c_p)$ is calculated for training:
\begin{equation}
\mathcal{L} = ||\epsilon_t - \epsilon_{\theta}(\bx_t,t,c_p)||_2^2.
\label{eqn:loss}
\end{equation}

\begin{figure*}[!t]
\centering
\includegraphics[width=1\linewidth]{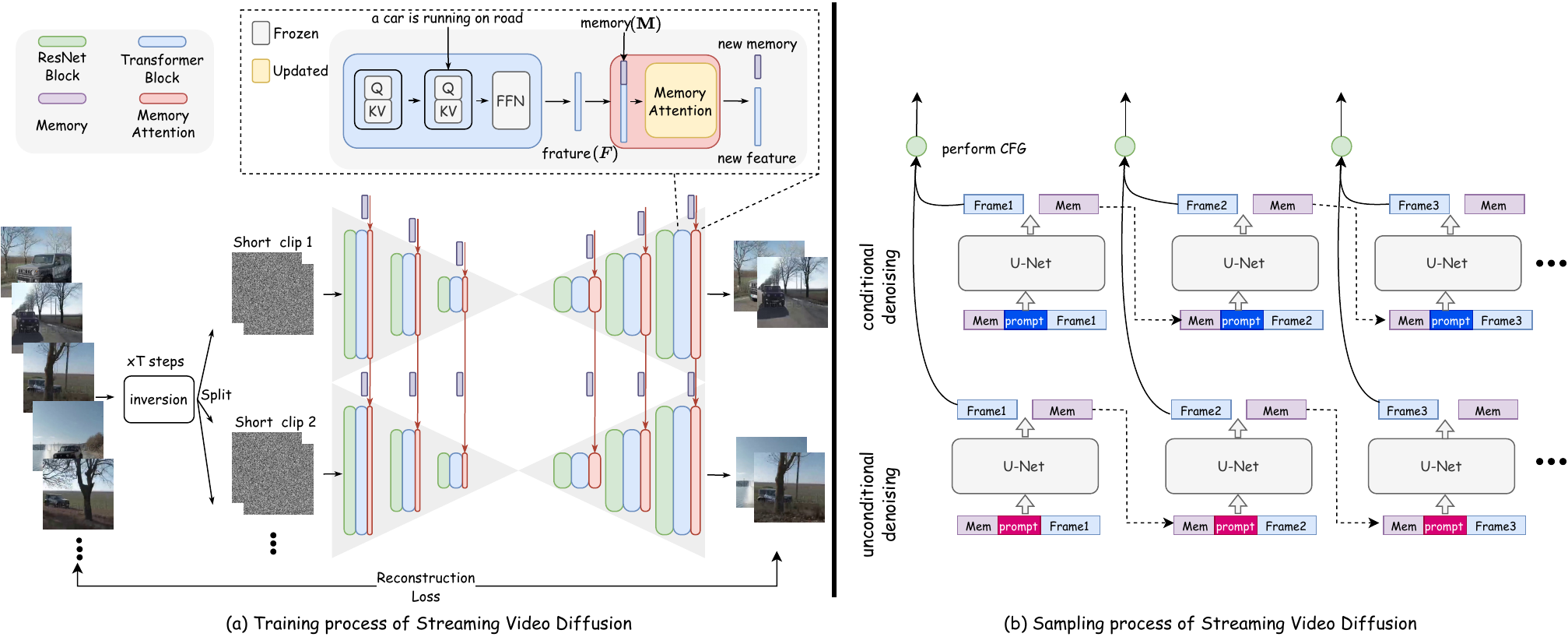}\caption{\label{fig:overview}Overview of Streaming Video Diffusion. (a) We propose spatial-aware temporal memory which is inserted with memory attention after each transformer block in Stable Diffusion. Then our method is trained on large-scale long videos by splitting the long video into short clips. (b) During inference, we denoise the noisy latent of streaming frame with classifier-free guidance (CFG) where each denoising step involves the U-Net conducting conditional and unconditional denoising with corresponding memory. %\bohan{add noise means inversion?? For inversion, you at least need to cite several references in the main text...}
}
\end{figure*}

\noindent To achieve fast editing, we obtain the edited latent representation $\hat{\bx}_0$, by sampling from the noise $\bx_T$, which is derived from $\bx_0$ through LCM Inversion~\cite{lcm}. This process employs LCM LoRA~\cite{lcm} for efficient few-step denoising. The final edited image, $I'$, is then generated by decoding $\hat{\bx}_0$ using the decoder $\mathcal{D}(\hat{\bx}_0)$.
For each denoising step, we apply classifier-free guidance (CFG)~\cite{cfg} to balance the fidelity and controllability  using the following linear combination of the conditional
and unconditional score estimates:
\begin{equation}\label{eqn: cfg}    
\hat{\epsilon}_t= (1+\lambda) \underbrace{\epsilon_{\theta}(\bx_t, t, c_p)}_{\text{conditional}} - \lambda \underbrace{\epsilon_{\theta}(\bx_t, t, \varnothing)}_{\text{unconditional}}, 
\end{equation}
where $\lambda$ is the coefficient factor and $\varnothing$ is the null text embedding.
%\bohan{cite editing inversion references...see zhen's paper}
% \subsection{Streaming Video Diffusion}
% \begin{figure}[!t]
% \centering
% \includegraphics[width=0.7\linewidth]{figure1.pdf}\caption{Overview of our streaming video diffusion.}
% \label{overview}
% \end{figure}

%\paragraph{Problem definition of online video editing.} 
%Given the $i$-th streaming frame $\mathcal{I}_i$ of video $\mathcal{I}$, online video editing aims to edit it according to the text prompt $c_p$ by considering the temporal relation with previous $i-1$ frames in an autoregressive manner. 

\section{Method}
 To balance the tradeoff between computational cost and long-range temporal modeling, we introduce a novel approach known as SVDiff for online video editing.
 In \cref{sec overview}, we first give an overview of the proposed approach. 
 In \cref{sec:Compact Recursive Temporal Memory}, we introduce spatial-aware temporal memory that is designed for online video editing. 
 Then, In \cref{sec:Efficient training and inference}, we elaborate the training and inference procedure.

 \subsection{Overview}\label{sec overview}
Our method aims to extend the T2I diffusion to T2V diffusion for online video editing by incorporating compact temporal recurrence.
 The overview of SVDiff is illustrated in \cref{fig:overview}. In \cref{fig:overview} (a), we first insert a learnable spatial-aware temporal memory with recurrent memory attention (as explained in \cref{sec:Compact Recursive Temporal Memory}) after each transformer block of Stable Diffusion.
% \bohan{change here..}
 This memory functions as a dynamic, temporal cache, constantly updated to capture the details of each video frame.
 Given a video $\mathcal{V} = \{\vec{V}^i\}_{i=1}^N$ with $N$ frames, we split it to $K$ short clips after inversion where $\mathcal{V} = \{ \vec{S}^i\}_{i=1}^K$. 
 Then we sequentially process each video clip with recursive spatial-aware temporal memory and calculate the reconstruction loss for training. 
 In this way, SVDiff can efficiently learn compact temporal memory over long videos. 
 During inference, as shown in 
 \cref{fig:overview} (b), each streaming frame $\vec{V}^i$ undergoes multiple denoising steps in which every step keeps a conditional and unconditional memory to maintain content and motion consistency.

\subsection{Spatial-aware Temporal Memory}
\label{sec:Compact Recursive Temporal Memory}

 We propose a spatial-aware temporal memory which is a learnable temporal embedding that recursively captures and updates temporal information from previous frames~\cite{bulatov2022recurrent}. 
We represent the temporal history until the $n$-th frame as a learnable memory embedding $\vec{M}^n \in \mathbb{R}^{h\times w \times d}$ where $h \times w$ is the spatial dimension and $d$ is the feature dimension. We note that simply increasing the size of this learnable memory does not inherently provide order and structural correlation with the spatial layout of frames, making it inadequate for capturing the motion trajectories of individual objects.
 %To enhance spatial awareness of memory, we configure $\mathbf{M}^n=[\vec{m}_{ij}]$ in $h \times w$ a grid  
Therefore, for the feature $\vec{m}_{ij}$ in position $(i, j)$ of $\vec{M}^n$, we augment it with the position embedding to enhance spatial awareness of memory, implicitly analogous to the spatial layout of the frame feature.
%where each grid $(i, j)$ consists of a latent feature $\vec{m}_{ij}$,  implicitly analogous to the spatial layout of the frame feature.
%Each grid $(i, j)$ consists of a latent feature $\vec{m}^n_{ij} \in \mathbb{R}^{d}$.
%However, simply increasing the size of this learnable memory does not inherently provide order and structural correlation with the spatial layout of frames, making it inadequate for capturing the motion trajectories of individual objects.
Following~~\cite{vit,chen2023object}, we add positional embedding in $\vec{M}^n$ where the position of $\vec{m}_{ij}$ is computed relatively to the center of the map as $(i-\frac{h}{2}, j-\frac{w}{2})$.  Therefore, such a memory shares a spatial structure similar to that of frame features.
%\bohan{it is not clear to me why the memory map is implicit. You learn an implicit mapping function like NeRF?? What is cell?}
In detail, at the beginning of a video in each timestep, we initialize $\vec{M}^0$ using the grid position:
\begin{equation}
\label{eqn:imap_init}
    \vec{m}_{ij} = \vec{w}^0 + \text{FFN}([i-\frac{h}{2}, j-\frac{w}{2}]),
\end{equation}
where $\vec{w}^0 \in \mathbb{R}^d$ is a learnable embedding and $\text{FFN}$ is a feed-forward network consisting of two MLP layers. Then, the frame features can be aligned with historical motion by temporal memory and recurrent memory attention~\cite{bulatov2022recurrent} as:
\begin{equation}
    [\vec{F}^n; \vec{M}^{n+1}] = \text{Attn}([\vec{F}^n; \vec{M}^n]),
    \label{eqn:attn}
\end{equation}
where $\vec{F}^n$ is the $n$-th frame features from the output of transformer block and $[\cdot;\cdot]$ is the concatenation operation. $\vec{M}^{n+1}$ is the updated memory for the next frame and $\Vec{M}^0 = [\vec{m}_{ij}] \in \mathbb{R}^{h \times w \times c}$. Memory attention $\text{Attn}(\cdot)$ is a standard self-attention module processing $[\vec{F}^n; \vec{M}^n]$ along the spatial dimension. 

Compared to explicit temporal memory collected from the key-value of previous frames~\cite{zhao2023streaming}, our spatial-aware temporal memory is more efficient in (1) condensing historical information in a compact memory and (2) learning temporal memory in different denoising time steps. This spatial-aware temporal memory is integrated into the original U-Net following each Transformer block.

\subsection{Efficient Training and Inference}
\label{sec:Efficient training and inference}

We train SVDiff on a large-scale, long video dataset~\cite{Bain21} to enable online video editing with extended streams. 
%\bohan{which dataset? any reference?}
Unlike existing methods~\cite{wu2022tuneavideo,ma2023follow} which are usually trained with 16-frame videos due to memory limits, we propose to solve this issue by splitting each long video into several short video clips. Given the $k$-th clip $\vec{S}^k$, we sequentially align each frame feature $\vec{F}^i$ in $\vec{S}^k$ with temporal memory by \cref{eqn:attn} and calculate the reconstruction loss using \cref{eqn:loss} between the predicted noise and the real noise latent of each frame.
Moving to next clip, we propagate memory from the output of the last frame of $\vec{S}^{k}$ to the beginning frame of $\vec{S}^{k+1}$. Therefore, the historical temporal information is still accessible in the following clips. 
 This process recursively continues until all frames are involved in the training.
%\bohan{cannot understand the last sentence..} 
In our method, we selectively update the learnable spatial-aware memory and its associated memory attention, as shown in \cref{fig:overview} (a). This is designed to enhance computational efficiency while preserving the original property of pre-trained T2I diffusion.
%\bohan{you need to use Figure 2 to figure out these modules...} 

During the inference stage, we first inverse the original video frame by frame into the noisy latent $\vec{x}_T$ and then use the classifier-free guidance (CFG)~\cite{cfg} to achieve denoising of the streaming frames.
%\bohan{it is better to mention CFG in preliminary..mention conditional/unconditional denoising..}
Specifically, as shown in \cref{fig:overview} (b), each denoising step involves the U-Net conducting two separate predictions: one for the conditional denoising and the other for the unconditional denoising, which are denoted by subscript $c$ and $uc$ respectively.
Therefore, we designate conditional and unconditional memory $\vec{M}_c$ and $\vec{M}_{uc}$ for them separately. Given the $n$-th frame, we can obtain estimated noise $\epsilon_t^n$ by modifying \cref{eqn: cfg} into:

\begin{equation}
\begin{aligned}
\hat{\epsilon}_{t,c}^{n}, \vec{M}^{n+1}_c &= \epsilon_{\theta}(\bx_t^n,t,c_p,\vec{M}_c^{n}), \\
\hat{\epsilon}_{t,uc}^{n}, \vec{M}^{n+1}_{uc} &= \epsilon_{\theta}(\bx_t^n,t,\varnothing,\vec{M}_{uc}^{n}), \\
\hat{\epsilon}_t^n =  (1 &+ \lambda) \hat{\epsilon}_{t,c}^n - \lambda \hat{\epsilon}_{t,uc}^n,
\end{aligned}
\end{equation}
where $\epsilon_{\theta}$ denotes the denoising U-Net.

% \begin{algorithm}[t]
% \caption{Training process of SVDiff}\label{alg:train}
% \begin{algorithmic}[1]
% \Procedure{DiffusionTraining}{$\mathcal{A}$}
%     \For{each epoch}
%         \State $\textbf{v}$ = Dataloader($\mathcal{A}$), $t \sim \text{Uniform}(0, T)$
%         \State Initialize $M$ using \cref{eqn:imap_init}
%         \For{each clip $s^k$ in $\textbf{v}$} 
%             \State Calculate noisy latent of $s^k$ as $\{\bx_t\}$ using \cref{eq:forward_sample}
%             \State $\mathcal{L}_{clip}^k, M = \Call{ClipProcess}{(\{\bx_t\}, t, c_p, M)}$  
%         \EndFor
%         \State Sum over $\mathcal{L}_{clip}$ as $\mathcal{L}_{video}$
%         \State Backpropagate and update $\epsilon_\theta$ with $\mathcal{L}_{video}$
%     \EndFor
%     \State \Return trained denoiser $\epsilon_\theta$
% \EndProcedure

% \Procedure{ClipProcess}{$\{\bx_t\}, t, c_p, M$}
%     \State Initialize index $i$ = 0
%     \For{$\bx_t$ in $\{\bx_t\}$}
%         \State $\epsilon, M_t = \epsilon_\theta(\bx_t, t, c_p, M_t)$
%         \State Calculate frame loss as $\mathcal{L}^i_{frame}$ using \cref{eqn:loss}
%         \State $ i \leftarrow i +1$
%     \EndFor
%     \State Sum over $\mathcal{L}_{frame}$ as $\mathcal{L}_{clip}$
%     \State \Return $\mathcal{L}_{clip}, M$
% \EndProcedure
% \end{algorithmic}
% \end{algorithm}

%% file: sec/4_result.tex
\section{Results}

\subsection{Experimental Settings}

%\bohan{use \noindent\textbf{xxxx}}
\noindent\textbf{Implementation details.} Our experiment is based on Stable Diffusion 1.5 with the public pretrained weights~\cite{rombach2022high}. 
We trained our model for 20k iterations using a subset of the HDVILA dataset~\cite{xue2022hdvila}, which comprises approximately 2 million subtitled videos. We sample 64 consecutive frames at a resolution of $512 \times 512$ from the input video for temporal consistency learning. 
To improve the efficiency of long video training, we divide the video into several video clips with 8 frames each. All clips are associated with the same video caption.
The training process is performed on 8 NVIDIA Tesla A100 GPUs and can be completed in eight days. 
For spatial-aware temporal memory, we empirically set $h = w = 8$. 
During inference, we utilize the LCM sampler combined with LCM LoRA~\cite{lcm} in 3 denoising steps to enhance inference speed. To further accelerate the infernce speed, we implement our method with TensorRT and
tiny AutoEncoder in StreamDiffusion~\cite{streamdiffusion}. Inference speed testing is conducted on an RTX 4090 GPU with images of $512 \times 512$ resolution.
Following~\cite{genlvideo}, we assess our approach on a dataset comprising 66 videos with lengths ranging from 32 to 150 frames. These videos are mostly drawn from the TGVE competition~\cite{tgve} and the Internet.

\noindent\textbf{Baseline models.}
There are generally four kinds of methods designed for online processing. We adopt these methods in online video editing as baseline models to verify the effectiveness of our method, including
\texttt{Efficient Attention}: We use efficient temporal attention with recurrent attention masks~\cite{katharopoulos2020transformers}.  
\texttt{Window Attention}: We cache the key-value of previous three frames for cross-frame interaction~\cite{zhao2023streaming,beltagy2020longformer}. 
%While efficient in inference, performance decreases dramatically once the keys and values are evicted. 
\texttt{Temporal Shift}: Following~\cite{liu2022ts2}, we inject temporal information by exchanging channel features with adjacent frames. %However, this temporal information is implicit and short-term. 
 \texttt{Sliding Window}: We use a fixed-length time window that incrementally moves over a video to analyze portions of it sequentially~\cite{ai2022class, transformerxl}. %This model collapses once the video length exceeds the cache size.  
%\bohan{this paragraph should not be put in the introduction..This is ablation study..also, any reference there?? They are quite common techniques used in the literature, both vision and NLP..} %We compared our approach with other four baseline models, including using efficient attention, window attention, temporal shift and sliding window. 
We elaborate the description and implementation details of these four baseline methods in Sec. 1 of supplementary.

\noindent\textbf{Evaluation metrics.} Following~\cite{wu2022tuneavideo}, we evaluate the performance of our method using CLIP metrics~\cite{clip} and user studies. To assess temporal consistency, we calculate CLIP image embeddings for all frames in the output videos and report the average cosine similarity between pairs of video frames.
%\bohan{CLIP reference?. you follow xxxx literature??}
For evaluating editing frame accuracy, we compute the average CLIP score between the frames of the output videos and their corresponding edited prompts.
In addition, we conduct three user study metrics, referred to as `Edit', `Image', and `Temp', to gauge the editing quality, overall frame-wise image fidelity, and temporal consistency of the videos, respectively. 
We conduct comparisons based on specific criteria between pairs of videos generated. In our user study, we enlist the feedback of 20 participants for each example and determine the final result based on majority voting.

%It's important to note that all results are assessed on videos with a minimum length of 100 frames, unless stated otherwise.

\begin{figure*}[!h]
\centering
\includegraphics[width=1\linewidth]{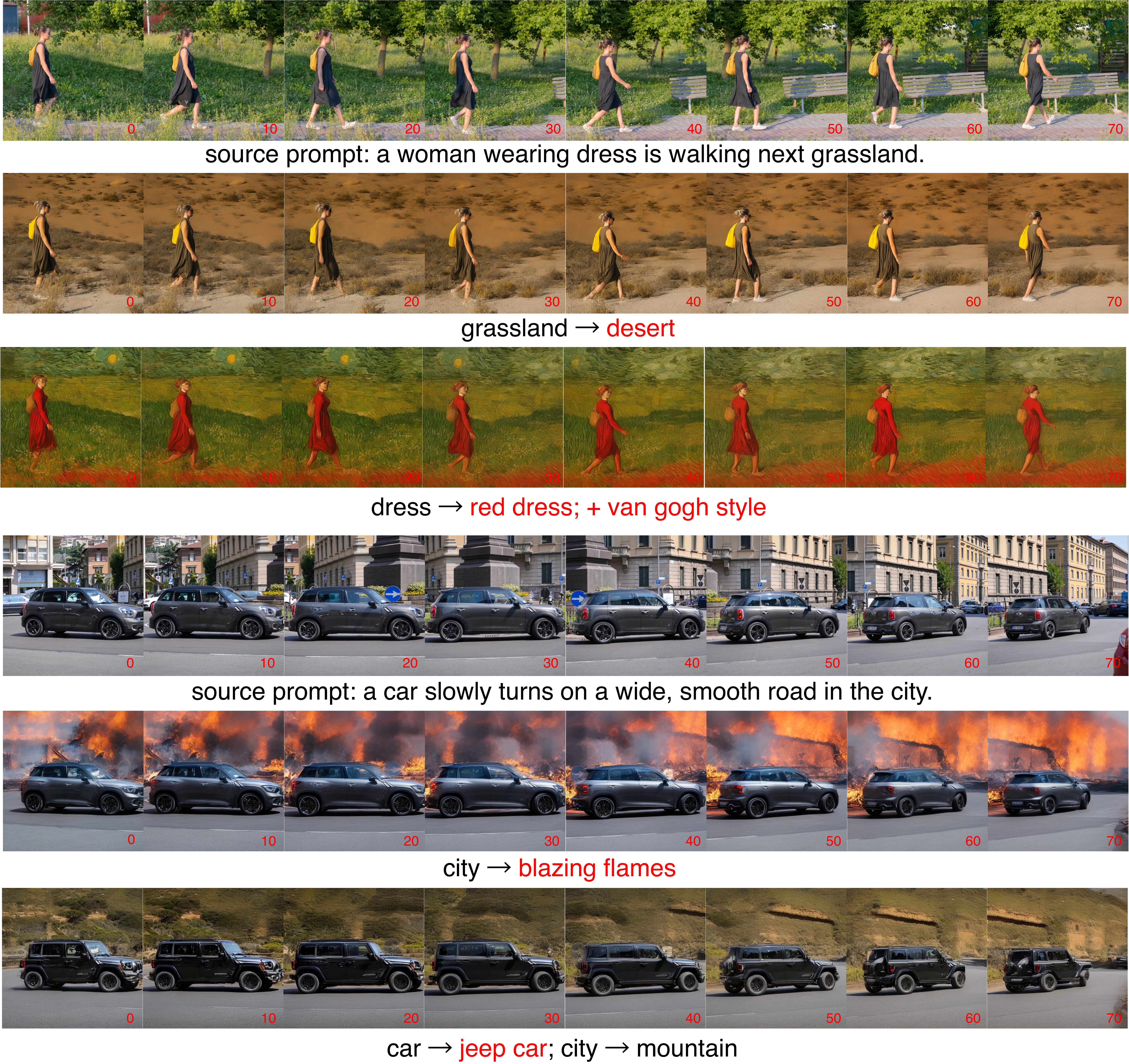}\caption{Qualitative editing results of long videos where the red number in the lower right corner denotes the frame index.} %\bohan{the image indexes are unclear..}}
\label{fig long}
\end{figure*}

\subsection{Editing Results}
 \cref{fig long} shows the application of our SVDiff on online video editing with extended video sequence. We note that it successfully produces high-quality videos closely aligned with the given text prompts. Specifically, SVDiff demonstrates proficiency in modifying global environments (\texttt{grassland $\rightarrow$ desert, city $\rightarrow$ blazing flams}), object attribute (\texttt{dress $\rightarrow$ red dress, car $\rightarrow$ jeep car}), and style (\texttt{van gogh style}). %Additionally, it can edit local object attributes, as shown in the \texttt{jeep car} featured in the seventh and eighth rows.
 It showcases the practical applicability of our method in the realm of online video editing. For more editing examples, please refer to the video in supplementary.

\begin{figure}[h]
\centering
\includegraphics[width=0.8\linewidth]{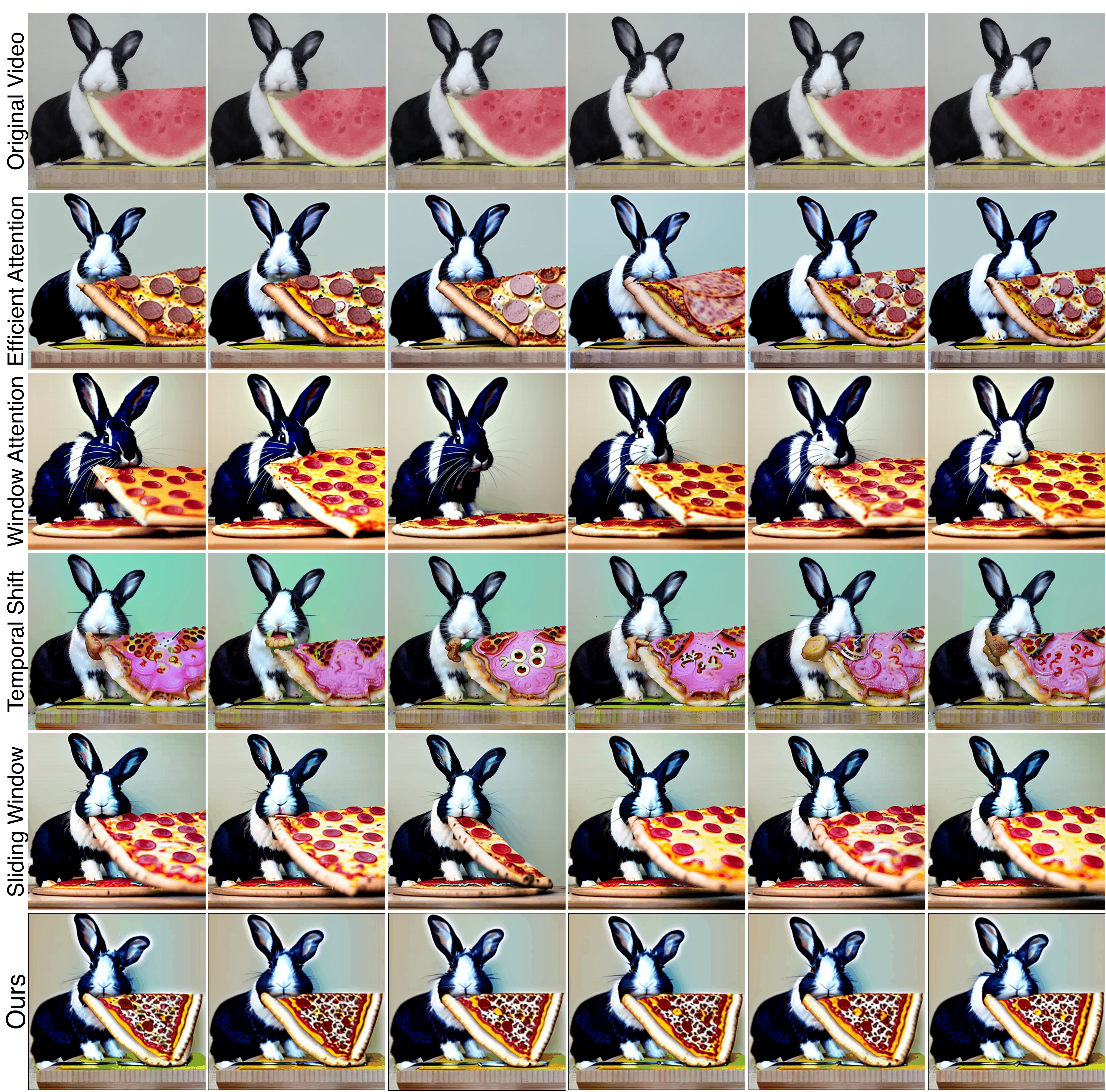}\caption{Visual comparison between baseline models and our method where the edit prompt is ``a rabbit is eating pizza". }
\label{figure online}
\end{figure}
\subsection{Comparison to Baseline Models}

\begin{table*}[!h]\caption{Quantitative comparison with other baseline models. We omit GFLOPs and FPS of efficient attention based method, since it is quadratic with the number of previous frames in streaming process.}
\centering
\scalebox{0.9}{
\begin{tabular}{lcccccccc}
\toprule
\multirow{2}{*}{Method} & \multicolumn{2}{c}{CLIP Metrics$\uparrow$} & \multicolumn{3}{c}{User Study$\downarrow$} & \multicolumn{3}{c}{Efficiency} \\
                        & Tem-Con        & Frame-Acc       & Edit     & Image     & Temp    &    GFLOPs            &    Params(MB)    & FPS       \\\hline
  Efficient Attention~\cite{katharopoulos2020transformers}                     &   90.87             &     27.34            &    4.35      &     3.79      &   2.60      &     -           &   344.7       & -     \\
Window Attention~\cite{zhao2023streaming}                     &  90.40             &     27.67            &  3.48     &  3.17         &   3.15      &         15520          &  344.7     &   12.3     \\
  Temporal Shift~\cite{liu2022ts2}                     &     91.67           &     27.56            &    2.70        &  3.42         &   4.37      &     12946           &   295.1     &  18.0\\ 
 Sliding Window\cite{transformerxl}                      &    90.82           &       27.37          &   2.64       &    2.85       &   3.10      &      46554         &   344.7   &  3.6 \\ 

   SVDiff(ours)                      &       \textbf{93.20}         &       \textbf{27.97}          &   \textbf{1.82}       &  \textbf{1.76}         &  \textbf{1.78}       &      15833          &   344.8    & 15.2 \\ 
  \bottomrule      
\end{tabular}}
\label{tab quantitative}
\end{table*}

\noindent\textbf{Quantitative results.} We provide quantitative comparisons with other baseline models in \cref{tab quantitative}.
Our SVDiff model demonstrates a notable improvement in the trade-off between performance and efficiency when compared to baseline models. Specifically, SVDiff outperforms the temporal shift-based method, achieving a 1.53\% improvement in temporal consistency with only extra 2,887 GFLOPs. Furthermore, the addition of 49.7MB in parameters, attributed to the incorporation of learnable memory embedding and memory attention in our model, is on par with the augmentations seen in efficient attention-based and sliding window-based approaches. Notably, our method requires significantly fewer GFLOPs and attains a real-time inference speed of 15.2 FPS. %which aligns with the performance of existing text-to-image diffusion methods(Temporal Shift).
This enhanced performance is attributed to SVDiff's recurrent temporal modeling, which efficiently integrates all previous temporal information into a spatial-aware compact memory. %These results collectively demonstrate the superior capability of SVDiff in balancing performance and computational efficiency. 

\noindent\textbf{Qualitative results.} We present a visual comparison in \cref{figure online} against baseline models to qualitatively assess the improvement of our method. Our method (bottom row) produces videos that better adhere to the edit prompt and preserve the temporal consistency of the edited video, while other methods struggle to meet both of these goals. For example, the efficient attention-based method has a sudden change in the \texttt{pizza} where the third and fourth figures of the first row have different appearances and shapes. 
Window attention and sliding window-based methods preserve temporal adherence between adjacent frames but suffer from long-term temporal inconsistency in the \texttt{rabbit} face and \texttt{pizza} texture. 
Moreover, the temporal shift-based method can maintain general style across frames, however, the temporal consistency in detailed texture is still poor (\texttt{pizza} in forth row).

\begin{figure}[t]
\centering
\includegraphics[width=0.5\linewidth]{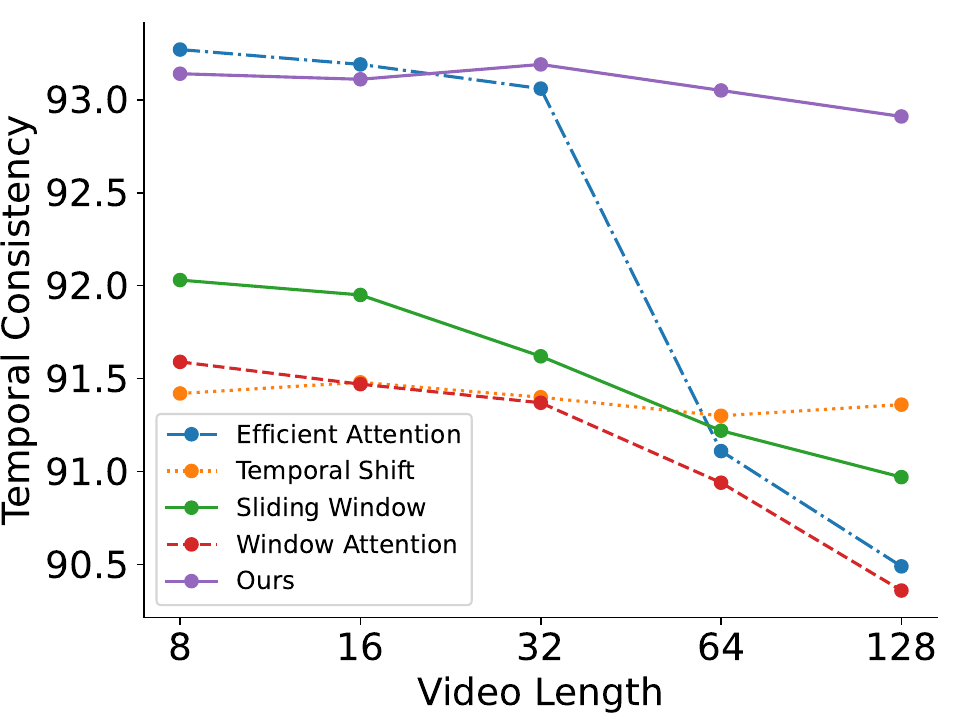}\caption{Performance comparison with baseline models in long video editing with different video lengths. %\bohan{enlarge the font size}
}
\label{figure length}
\end{figure}
\noindent\textbf{Video editing with different lengths.} We test the performance of video editing with different lengths in \cref{figure length}. We observe that efficient attention, window attention, and sliding window based methods meet various degrees of performance decline in editing long videos. However, the reasons are different. Specifically, the efficient attention-based method is limited by the training-inference gap where once the video length of inference is larger than that of training, the performance decreases sharply. For window attention and sliding window-based methods, they are largely influenced by the window size that accesses previous frames. Besides, the temporal shift-based method is stable in editing videos of different lengths, but its training-free temporal modeling strategy can only provide implicit nearby temporal information. In addition, our method achieves superior performance over them because of our recursive spatial-aware temporal memory and training with longer videos.

\subsection{Comparison to Existing Models}
\begin{figure}[h]
\centering
\includegraphics[width=0.8\linewidth]{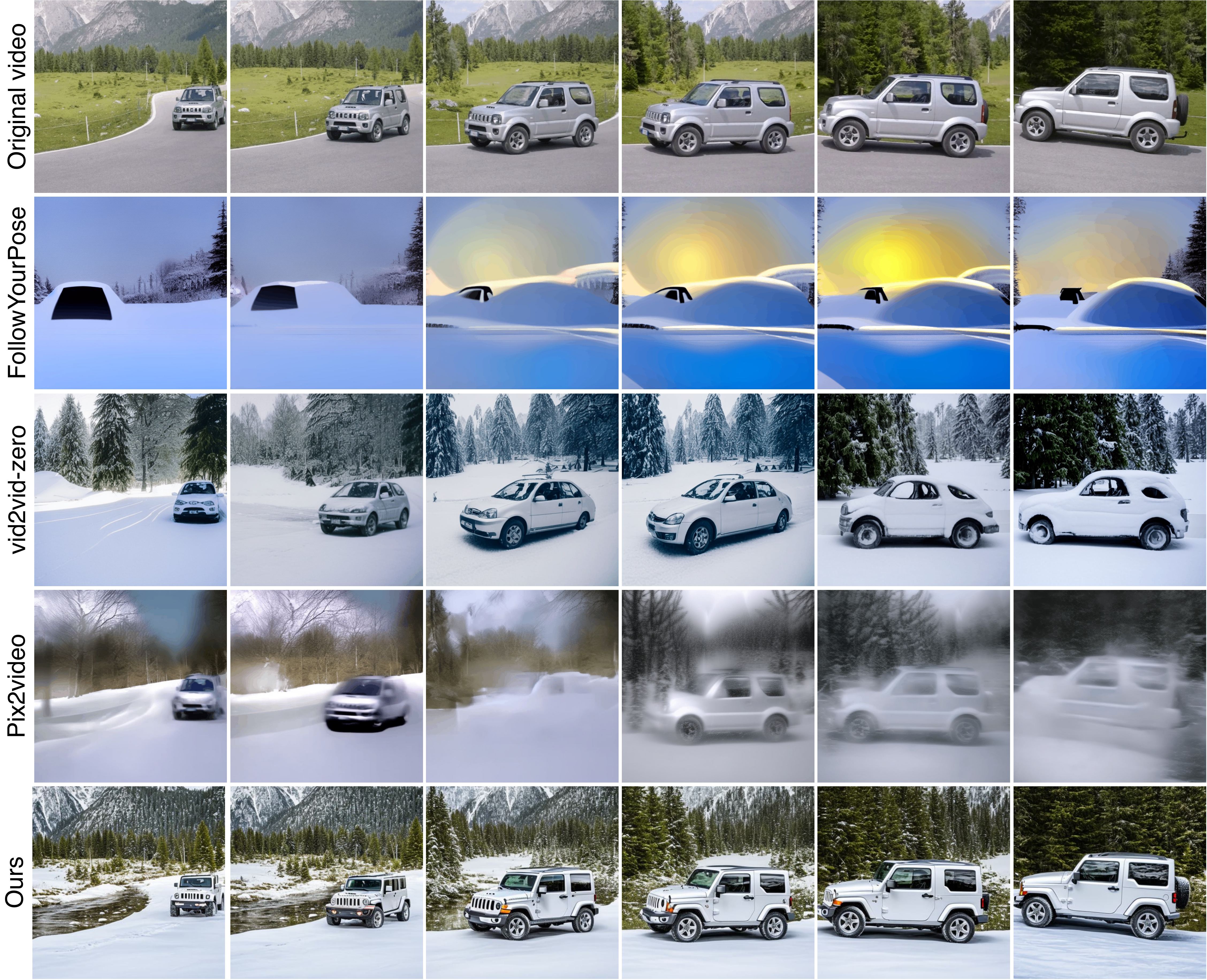}\caption{Visual comparison with existing methods that adapt to online video edit where the edit prompt is ``a car is moving on snow road". }
\label{comparison}
\end{figure}

\noindent\textbf{Quantitative results.} As discussed in \cref{sec:intro}, existing tuning-free and pretrained-based methods can be adapted to online video editing. 
%\bohan{what kind of methods? can you make it specific?}
Therefore, we compare the modified pre-trained method (FollowYourPose~\cite{ma2023follow}) and tuning-free based methods (vid2vid-zero~\cite{wang2018vid2vid} and Pix2video~\cite{ceylan2023pix2video}) with our method on online video editing in \cref{tab offline}. The implementation details of these methods are in Sec. 1 of supplementary. Our method demonstrates a significant edge over the other three in terms of CLIP metrics and user study. Specifically, it surpasses FollowYourPose by 3.49\% in temporal consistency and by 2.15 in editing quality. This improvement is attributed to the consistent recurrent operations of our method applied to long videos during both training and inference. Moreover, our method outperforms Pix2video and vid2vid-zero as well. We ascertain that their reliance on adjacent frames for sparse causal attention falls short in modeling long-term motion trajectories.
%\bohan{at least one sentence of results analysis }

\noindent\textbf{Qualitative results.}  We present a visual comparison in \cref{comparison} against existing methods that are adapted to online video editing to assess the improvement of our method. Our method, depicted in the bottom row, excels in adhering to the edit prompt while maintaining the temporal consistency of the edited video. In contrast, FollowYourPose~\cite{ma2023follow} exhibits notable challenges in preserving original motion and content integrity. Pix2video~\cite{ceylan2023pix2video} tends to generate visuals of inferior quality, marked by blurriness and inconsistencies in object continuity. Additionally, vid2vid-zero~\cite{wang2018vid2vid} demonstrates a clear disparity, particularly in the representation of a \texttt{car}: While the final image features a snow-covered vehicle, the \texttt{car} appears clean in the preceding images. These comparisons underscore our method's unique ability to keep edit adherence and temporal consistency, outperforming existing approaches in online video editing scenarios.

\begin{table}[t]
\caption{Quantitative comparison with existing methods that adapt to online video editing. %\bohan{we are the first one? add reference to the compared method...}
}
\centering
\footnotesize{
\begin{adjustbox}{width=0.7\textwidth,center}
\begin{tabular}{lccccc}
\toprule
\multirow{2}{*}{Method} & \multicolumn{2}{c}{CLIP Metrics$\uparrow$} & \multicolumn{3}{c}{User Study$\downarrow$}  \\
                        & Tem-Con        & Frame-Acc       & Edit     & Image     & Temp             \\\hline
 FollowYourPose~\cite{ma2023follow}                    &  89.71             &    26.70           &    3.70      &  3.52         &     3.28           \\
 vid2vid-zero~\cite{wang2018vid2vid}                  &  91.68           &  27.88             &    2.66      &    2.05       &     1.70       \\
 Pix2video~\cite{ceylan2023pix2video}                    &  91.27             &   27.61          &   2.10      &   3.02      & 3.62      \\ 
 SVDiff(ours)                      &   \textbf{93.20}         &   \textbf{27.97}           &    \textbf{1.55}      &   \textbf{1.41}        &  \textbf{1.40}         \\ 
\bottomrule      
\end{tabular}
\end{adjustbox}
}
\label{tab offline}
\end{table}

\subsection{Ablation Study}

\begin{figure}[tbp]
\centering
\includegraphics[width=0.8\linewidth]{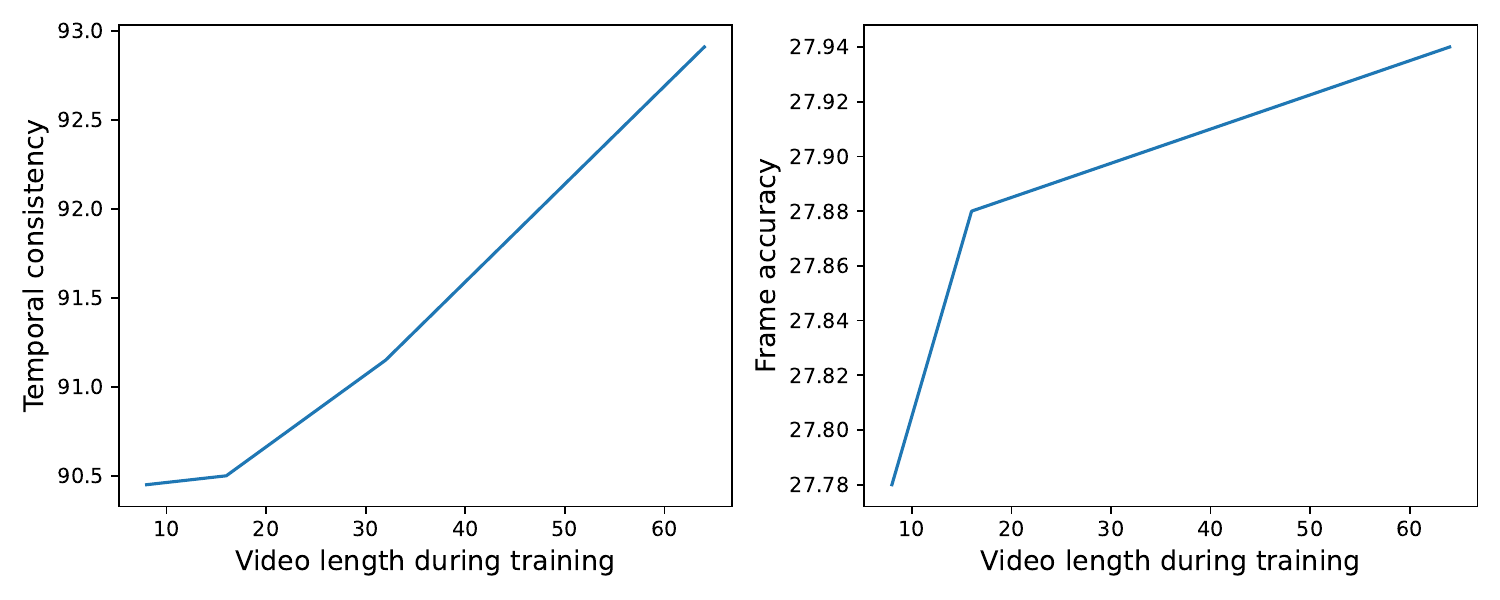}\caption{Ablation study on video length during training.}
\label{figure training_frame}
\end{figure}

\textbf{Training with longer videos.} One benefit of our SVDiff is training on longer videos. In \cref{figure training_frame}, we ablate the influence of video length during training. By increasing the length of the video from 8 to 64, the temporal consistency increases monotonically with a gain of 2.5\%, indicating the benefit of training on long videos. However, for frame accuracy, such improvement becomes negligible where largely increasing video length only brings a gain of 0.15\%. This is because the temporal module trained on long videos can effectively learn temporal coherence, but the editing ability is largely determined by the base model which is frozen during training.%This is because although training on longer videos can directly alleviate the training-inference frame gap for temporal consistency, it can not inherently enhance the model's ability to align individual frames with specific editing instructions.
%\bohan{hard to understand the last sentence..}

\begin{figure}[tbp]
\centering
\includegraphics[width=0.8\linewidth]{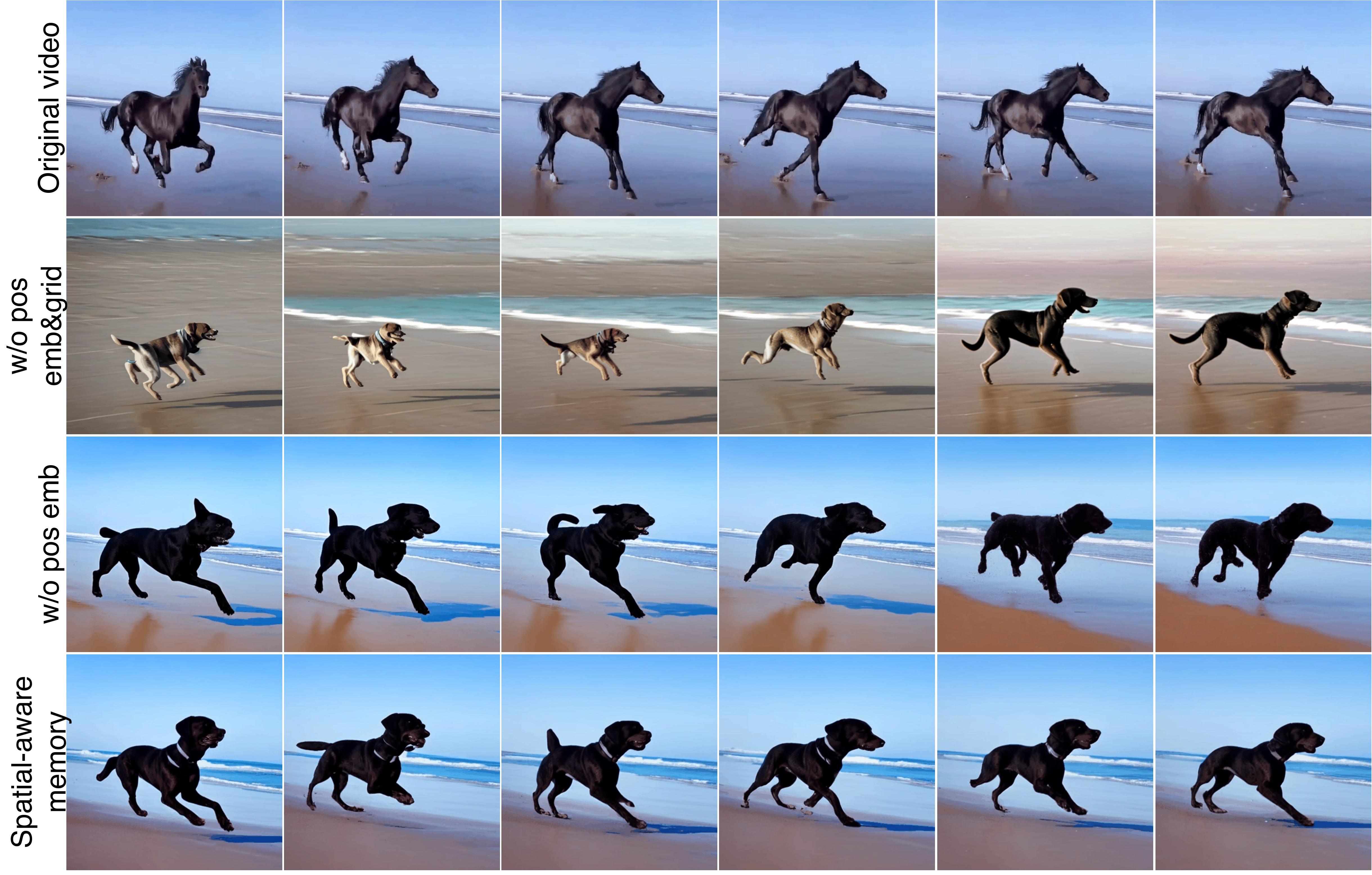}\caption{Ablation on spatial-aware temporal memory. The edit prompt is ``a dog is running on the beach". }
\label{figure ablation}
\end{figure}

\noindent\textbf{Spatial-aware temporal memory.} The effectiveness of our proposed spatial-aware memory is analyzed in \cref{figure ablation}. We observe that employing spatial-aware memory retains the intricate skeleton of the \texttt{horse}. In contrast, omitting positional embedding leads to noticeable losses, such as the disappearance of the \texttt{dog}'s legs in the third row of \cref{figure ablation}. Further, the removal of both positional embedding and grid format, akin to using a global token, results in the model maintaining only the broad content and style across frames. This causes inconsistencies like fluctuating \texttt{dog} sizes and shifting \texttt{sea} positions. These observations demonstrate the critical role of our proposed memory in preserving the detailed spatial layout of each frame and the inter-frame motion trajectory within the video stream.%\textcolor{blue}{changing showcase figure}%``Global Memory" denotes we configure the memory with a memory token while ``Local Memory" is the grid-format memory without adding position embedding. When using a global memory token to temporal information,
\begin{table}[h]\caption{Ablation on memory size of spatial-aware temporal memory.}
\centering
\begin{tabular}{ccc}
\toprule
Memory size & Tem-Con & Frame-Acc \\
\midrule
1$\times$1           &    89.96     &    26.30       \\
8 $\times$8          &  \textbf{92.91}       &   \textbf{27.94}        \\
16 $\times$ 16         &    92.70     &    27.59       \\ \bottomrule 
\end{tabular}\label{tab ablation memory}
\end{table}

Moreover, in \cref{tab ablation memory}, we ablate the memory size of our spatial-aware temporal memory $\vec{M}^n$. A smaller memory size (1$\times$1) lacks the granularity needed to effectively model spatial variations and temporal transitions, resulting in lower performance. Conversely, a larger memory size (16$\times$16) introduces redundancy and potential overfitting to specific frame details.

%% file: sec/5_conclusion.tex
\section{Conclusion}

In this paper, we have presented a new task called online video editing, which is designed to edit streaming frames while preserving temporal consistency. To this end, we have proposed Streaming Video Diffusion (SVDiff) to address the three challenges of this task by integrating compact spatial-aware temporal recurrence into existing Stable Diffusion. To train our method on long videos, we divide the video into short clips while preserving the long-term temporal coherence with the help of a compact temporal recurrence module. The experiments show that our SVDiff produces high-quality long videos with both global and local coherence and reduces the computation cost for streaming processing compared to baseline methods.
%\bohan{reduces...compare with what?}

\noindent\textbf{Limitations and future work.}  
Although in theory we can process videos of any length using temporal recurrence, our current method may not be able to accurately detect shot changes in videos longer than 2 minutes with thousands of frames, particularly those with discontinuous backgrounds and complex motion. This limitation largely stems from the gap between training and inference of video frames.
%\bohan{of video frames?}
Therefore, our future work will focus on investigating strategies to alleviate this influence, with the goal of efficiently processing long videos that feature complex scene transitions. %\bohan{why multi-step denoising and U-Net structure can increase memory?} %This is a compromise we have to make in order to achieve fine-grained temporal propagation and efficient memory storage for online video editing, as the cost of memory will be greatly increased by multi-step denoising and U-Net structure. \bohan{why multi-step denoising and U-Net structure can increase memory?}

%\bohan{line space wrong}